\documentclass[letterpaper]{article}

\usepackage{aaai2027}

\usepackage[hyphens]{url}
\usepackage{graphicx}
\usepackage{natbib}
\usepackage{caption}
\usepackage{algorithm}
\usepackage{algorithmic}
\usepackage{makecell}
\usepackage{newfloat}
\usepackage{listings}
\usepackage{booktabs}
\usepackage{multirow}
\usepackage{amsmath}
\usepackage{amssymb}
\DeclareCaptionStyle{ruled}{
    labelfont=normalfont,
    labelsep=colon,
    strut=off
}

\floatstyle{ruled}
\newfloat{listing}{tb}{lst}{}
\floatname{listing}{Listing}

\title{Who Leads Now? Token-Level Modality Arbitration for Chart-to-Code Generation}

\author{
Qinghao Fu\textsuperscript{\rm 1,\rm 2}\equalcontrib,
Yarong Wang\textsuperscript{\rm 1,\rm 2}\equalcontrib,
Shunlei Ning\textsuperscript{\rm 2}\equalcontrib,
Yilin Wang\textsuperscript{\rm 1},
Shunwen Bai\textsuperscript{\rm 1,\rm 2},\\
Xinda Wang\textsuperscript{\rm 3},
Jiaotuan Wang\textsuperscript{\rm 2},
Yinan Nie\textsuperscript{\rm 4},
Wei Zhou\textsuperscript{\rm 2}\corresponding
}

\affiliations{
\textsuperscript{\rm 1}Zhejiang University,
\textsuperscript{\rm 2}Ant Group,
\textsuperscript{\rm 3}Peking University,
\textsuperscript{\rm 4}Fudan University
}

\begin{document}
\makeatletter
\renewcommand{\copyright@text}{}
\makeatother
\maketitle

\begin{abstract}
Chart-to-code generation requires a model to read the fine-grained visual details of a chart and write executable code that reproduces it. Existing chart-to-code methods either train visual and coding abilities separately, or fine-tune on chart-to-code data with the two abilities entangled. Neither strategy accounts for the distinct nature of the two abilities or the interference that arises when they are optimized together. We propose MoCA (Mixture of Cross-modal Arbitration), which separates the two abilities rather than blending them. MoCA is built on Cross-modal Arbitration Block (CAB), which maintains a visual branch and a code branch as two distinct pathways, and a lightweight arbiter that arbitrates their relative contributions at every layer and generated token. We train MoCA in two stages: a supervised warm-up on self-distilled reasoning trajectories that decomposes visual understanding into explicit steps, followed by reinforcement learning with rewards on both the reasoning process and the final code. Analysis shows that the arbiter learns structured rather than arbitrary allocations, with expert contributions varying systematically across tokens, layers, and instances. Across three benchmarks, MoCA delivers competitive performance against general-domain and chart-specialized models. Ablation results show that the gains cannot be attributed to a larger model size alone, but instead arise from the joint contributions of complementary visual and code branch initialization and input-conditioned arbitration through CAB.

\end{abstract}

\section{Introduction}
\begin{figure}[t]
    \centering
    \includegraphics[width=\linewidth]{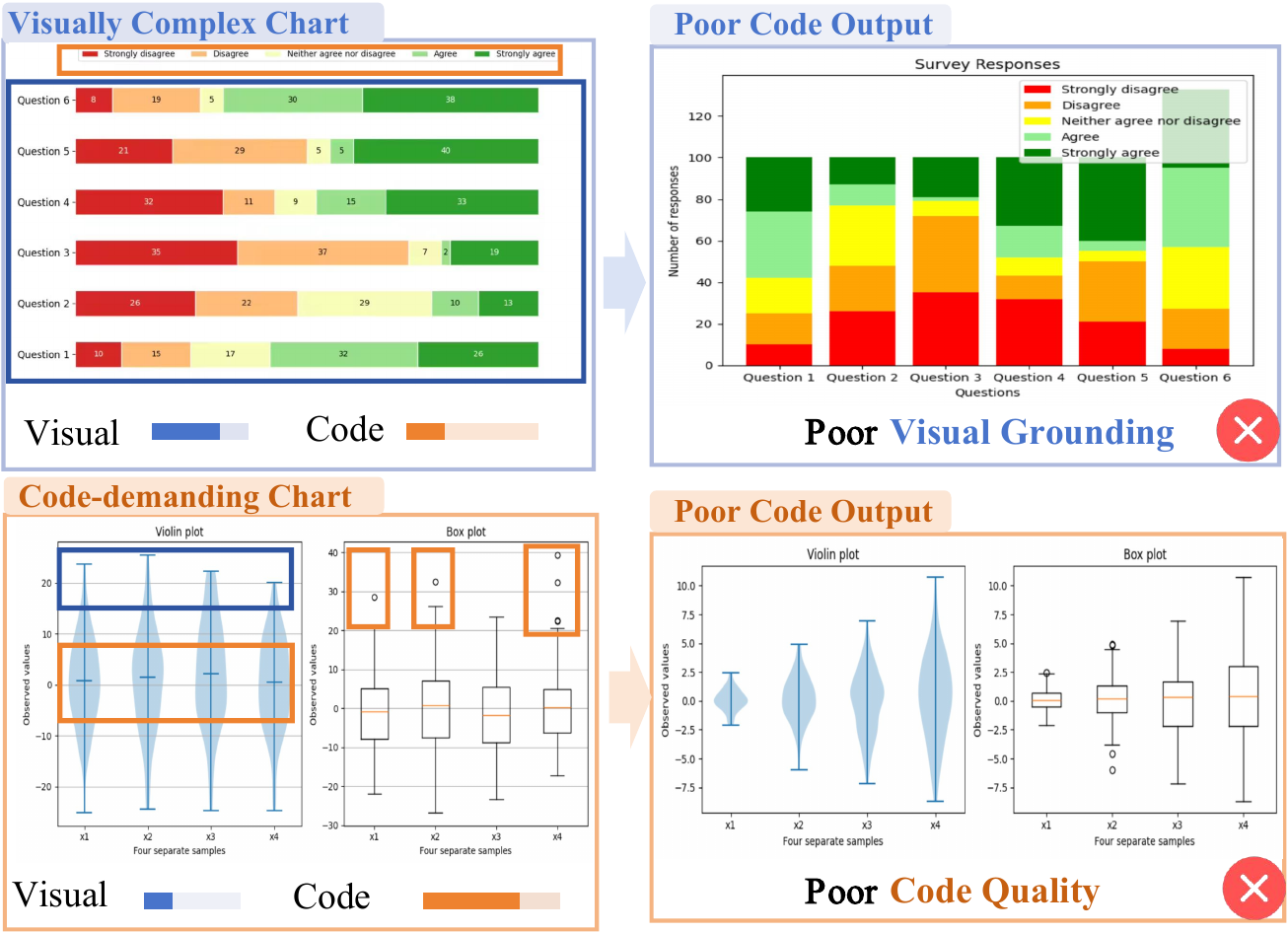}
   \caption{\textbf{Motivation of our work.} Chart-to-code generation places unequal demands on visual grounding and code generation across inputs. \textbf{\textit{Top:}} a visually complex chart, where the model misreads dense labels, subtle colors, and multi-component layouts during visual understanding. \textbf{\textit{Bottom:}} a code-demanding chart, where the model correctly perceives the chart but fails to generate the plotting logic such as stacked structures and custom axes. }
    \label{fig:intro}
\end{figure}

Charts are widely used to present structured data in a compact and interpretable form. Recent advances in multimodal large language models (MLLMs) have enabled research on chart-to-code generation~\cite{tang2025charts}. The goal is to reconstruct a chart image as executable plotting code. Compared with chart understanding tasks such as captioning or question answering, this task is more challenging, since the output must be both syntactically correct and visually faithful. It must recover data, layout, text, color, and style from the input chart.

A key challenge in chart-to-code generation is that visual grounding and code generation are not uniformly required across different inputs. As shown in Figure~\ref{fig:intro}, some charts are visually complex, requiring accurate recognition of dense labels, subtle color variations, and multi-component layouts. Other charts are more code-demanding, where the main difficulty lies in generating correct plotting logic such as stacked structures, reference lines, or customized axes. This suggests that the relative importance of visual and coding abilities varies across different inputs, rather than being fixed. In addition, different intermediate representations of the same input may emphasize different aspects of the task, such as visual alignment, structural reasoning, or execution correctness. However, existing chart-to-code methods typically treat visual understanding and code generation as either implicitly coupled capabilities or independently optimized components, without explicitly modeling their interaction during generation.

Existing methods improve performance from different directions, but they do not explicitly model this adaptive coordination problem. Strong multimodal models still struggle to generate correct and visually faithful chart code~\cite{zhao2025chartcoder,tang2025charts}. Training-based methods such as ChartMaster~\cite{tan2025chartmaster} improve performance through supervised fine-tuning (SFT) and reinforcement learning (RL), but the interaction between visual and coding abilities remains implicitly determined by the model. Agent-based methods such as MatPlotAgent~\cite{yang2024matplotagent} introduce iterative refinement with external feedback, but at the cost of increased inference complexity. Model-merging methods such as VisCodex~\cite{jiang2025viscodex} integrate vision and code capabilities into a single backbone using globally fixed parameter-space coefficients shared across all inputs.

In this work, we study chart-to-code generation from the perspective of adaptive capability coordination. We propose MoCA, a framework that maintains a visual branch and a code branch while dynamically coordinating their contributions during generation. At its core, CAB employs a lightweight Arbiter to predict branch weights from intermediate hidden representations for each token at every layer, thereby producing instance-specific capability allocation. This design enables MoCA to adapt the relative contributions of visual grounding and code generation according to both the input and the evolving generation context.

We train MoCA using a two-stage framework. During SFT, the model learns structured mappings from visual inputs to executable code using paired chart-code data and self-distilled Snippet-of-Thought (SoT) trajectories, establishing initial alignment between visual understanding and code generation. During RL, we optimize the model with multi-dimensional rewards covering format correctness, reasoning consistency, structural fidelity, execution validity, and visual similarity. Together, these rewards provide complementary process-level and outcome-level supervision to improve both code executability and visual fidelity.

We evaluate MoCA on three benchmarks: ChartMimic~\cite{yang2024chartmimic}, Plot2Code~\cite{wu2025plot2code}, and ChartX~\cite{xia2025chartx}. Across these benchmarks, MoCA delivers competitive performance among general-domain models and chart-specialized models. Controlled ablations against static merging and parameter-matched variants further suggest that the gains cannot be explained by model size alone, highlighting the importance of coordination between the visual and code branches. In summary, our contributions are three-fold:
\begin{itemize}
    \item We formulate chart-to-code generation from the perspective of cross-modal capability coordination, showing that visual grounding and code generation are complementary but heterogeneous abilities whose interaction should be explicitly modeled.
    
    \item We propose MoCA, a dual-branch architecture with CAB, which preserves visual and code expertise separately and employs a lightweight Arbiter to dynamically allocate their contributions based on hidden representations.
    
    \item We establish an effective training and analysis framework combining self-distilled SoT supervision with multi-dimensional reinforcement learning rewards, and provide empirical evidence that CAB learns structured cross-modal allocation patterns across tokens, layers, and instances.
\end{itemize}

\section{Related Work}
\textbf{MLLMs.}
MLLMs extend LLMs with visual understanding, enabling joint reasoning over images and text for a wide range of multimodal tasks~\cite{zhao2025chartcoder,tan2025chartmaster,zeng2024timesuite,kuang2025natural}. Early work aligns visual features with language representations through cross-modal projection or query-based interaction mechanisms~\cite{alayrac2022flamingo,li2023blip}. Building on this paradigm, recent models further introduce multimodal instruction tuning so that the model can follow natural language instructions grounded in visual inputs~\cite{zhu2023minigpt,liu2023visual}. These models have demonstrated strong performance on tasks such as visual question answering, document understanding, and diagram reasoning~\cite{kim2025visual,liao2025doclayllm,tan2025chartmaster}. However, generating structured outputs such as executable scripts remains challenging for general-domain MLLMs, since such tasks require not only visual grounding but also strong code generation ability~\cite{gui2025webcode2m,yang2025ui2code}. 

\textbf{Model Merging.}
Model merging aims to integrate capabilities from multiple pretrained models through parameter-space combination, typically without jointly retraining the source models or accessing their original training data~\cite{yang2026model}. The merged parameters can be deployed directly or used as initialization for subsequent task-specific adaptation. A basic approach is weight averaging~\cite{wortsman2022modelsoupsaveragingweights}, which ensembles knowledge in parameter space and improves generalization. Task arithmetic further shows that model behaviors can be composed or removed via linear operations on task vectors~\cite{ilharco2023editingmodelstaskarithmetic}.

Recent studies extend model merging along two main directions. The first focuses on improving merging quality through parameter-level or layer-level weighting~\cite{yang2024adamergingadaptivemodelmerging}. The second explores subspace or low-rank decomposition to reduce interference between tasks~\cite{yadav2023tiesmergingresolvinginterferencemerging,yu2024languagemodelssupermario}. In addition, some methods learn merging coefficients in an input-dependent manner~\cite{tang2024mergingmultitaskmodelsweightensembling,ye2025dynamic}. However, these approaches are primarily designed for single-modality or homogeneous task settings, and are less effective in cross-modal scenarios where different experts exhibit fundamentally different functional roles. Recent work such as VisCodex~\cite{jiang2025viscodex} explores merging vision-language and code models for multimodal generation, but relies on static merging weights and cannot adapt to varying visual and coding demands across inputs. In contrast, our work preserves the two capabilities as separate branches and introduces input-conditioned arbitration to dynamically coordinate their contributions during generation.

\textbf{Chart-to-Code Generation.}
Chart-to-code generation has recently attracted increasing attention as a task that requires models to generate executable plotting code grounded in fine-grained chart understanding. Early studies mainly focus on constructing datasets and benchmarks to evaluate model performance in this setting~\cite{tang2025charts,zhao2025chartedit}. Subsequent work improves generation quality by performing SFT on curated chart-code pairs and introducing structured intermediate supervision to guide the generation process~\cite{zhao2025chartcoder}. Recent methods further explore preference learning and reinforcement learning with textual, structural, execution-based, or visual rewards to improve code executability and chart reconstruction quality~\cite{zhang2025boostingcharttocodegenerationmllm}. These approaches demonstrate the importance of task-specific supervision for chart-to-code generation. However, they primarily improve the training signal or feedback objective, while the internal coordination between visual grounding and code synthesis remains underexplored. Our work addresses this gap by maintaining visual and code expertise as separate pathways and explicitly coordinating their interaction through a cross-modal arbitration mechanism, enabling adaptive capability usage during chart-to-code generation.

\begin{figure*}[t]
    \centering
\includegraphics[width=\textwidth]{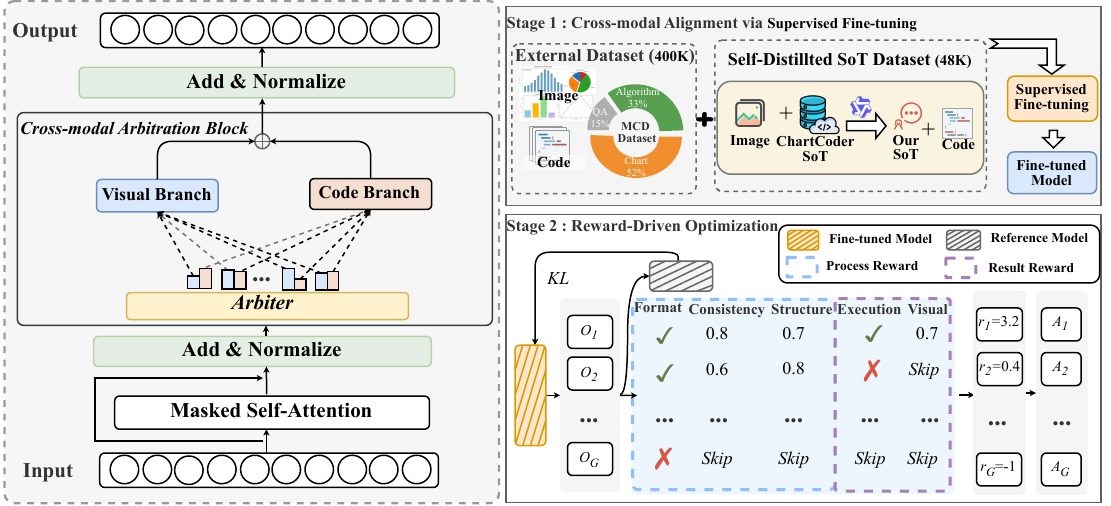}
    \caption{Overview of MoCA. The model incorporates CAB, where a lightweight, input-conditioned arbiter dynamically coordinates the visual and code branches. Training proceeds in two stages: Stage 1 performs SFT on external data and self-distilled SoT trajectories, while Stage 2 applies RL with multi-dimensional rewards.}
    \label{fig:model}
\end{figure*}

\section{Methodology}
We propose MoCA, a framework for chart-to-code generation that dynamically coordinates the visual and code branches based on the input and intermediate hidden representations. An overview of MoCA is shown in Figure~\ref{fig:model}.

\subsection{Problem Setting}
Let $\mathcal{D}=\{(I_i,x_i,C_i^\ast)\}_{i=1}^{N}$ denote a chart-to-code dataset, where $I_i$ is a chart image, $x_i$ is a natural-language instruction, and $C_i^\ast$ is the corresponding ground-truth program. The goal is to learn a model $\mathcal{M}$ that generates an executable program $C_i=(y_1,\ldots,y_T)$.

We formulate the autoregressive generation process as
\begin{equation}
P(C_i \mid I_i,x_i;\Theta)
=
\prod_{t=1}^{T}
P(y_t \mid y_{<t},I_i,x_i;\Theta).
\end{equation}

A key challenge is that visual grounding and code generation are required to different extents across chart structures and generation contexts. Some inputs demand fine-grained visual alignment, whereas others place greater emphasis on structural and syntactic code construction. Existing approaches generally combine these capabilities either implicitly within a single backbone or through globally fixed parameter-space coefficients, without explicitly adjusting their relative contributions across inputs and intermediate representations.

To address this limitation, MoCA dynamically predicts branch weights
$\boldsymbol{\alpha}_{i,l,t}
=(\alpha_{i,l,t}^{\mathrm{vis}},\alpha_{i,l,t}^{\mathrm{code}})$
from the hidden representation $\mathbf{h}_{i,l,t}$ for each token $t$ at layer $l$. These input-conditioned weights enable coordination between the visual and code branches throughout the generation process.

\subsection{Cross-Modal Arbitration}

To accommodate the heterogeneous requirements of chart-to-code generation, we introduce an input-conditioned coordination mechanism between a visual branch and a code branch. Unlike static parameter merging, which uses the same combination coefficients for all inputs, MoCA preserves the two branch transformations separately and dynamically adjusts their relative contributions based on intermediate hidden representations. This formulation enables fine-grained arbitration across inputs, layers, and generation steps.

\subsubsection{Branch Initialization}

MoCA uses Qwen2.5-VL-7B~\cite{qwen2025qwen25technicalreport} as its multimodal backbone and Qwen2.5-Coder-7B~\cite{hui2024qwen2} to initialize the code branch. Specifically, within each CAB, the visual branch is initialized from the feed-forward network parameters of Qwen2.5-VL-7B, while the code branch is initialized from the corresponding parameters of Qwen2.5-Coder-7B. The remaining backbone components retain their Qwen2.5-VL-7B initialization. This construction preserves the visual grounding capabilities of the multimodal backbone while introducing a prior for code synthesis.

\subsubsection{CAB}

To explicitly model the interaction between visual grounding and code generation, we introduce the CAB. CAB maintains a visual branch and a code branch as separate transformations, preserving their distinct functional roles. A lightweight Arbiter dynamically determines their relative contributions based on the current hidden representation. Consequently, the branch weights can vary across input instances, network layers, and generation tokens. Each CAB consists of three components:

\begin{itemize}
    \item \textbf{Arbiter:} a lightweight network that predicts the relative weights of the two branches from the current hidden representation.
    \item \textbf{Visual branch ($\mathcal{B}_{\mathrm{vis}}$):} a feed-forward transformation initialized from the corresponding FFN parameters of the vision-language model.
    \item \textbf{Code branch ($\mathcal{B}_{\mathrm{code}}$):} a feed-forward transformation initialized from the corresponding FFN parameters of the code model.
\end{itemize}

For sample $i$ and token $t$ at layer $l$, the Arbiter computes the branch weights as follows:
\begin{equation}
\boldsymbol{\alpha}_{i,l,t}
=
\operatorname{Softmax}
\left(
\mathbf{W}_{a}^{(l)}\mathbf{h}_{i,l,t}
+
\mathbf{b}_{a}^{(l)}
\right),
\end{equation}
where
$\boldsymbol{\alpha}_{i,l,t}
=
(\alpha_{i,l,t}^{\mathrm{vis}},
\alpha_{i,l,t}^{\mathrm{code}})$. The output of CAB is then computed as
\begin{equation}
\mathbf{z}_{i,l,t}
=
\alpha_{i,l,t}^{\mathrm{vis}}
\mathcal{B}_{\mathrm{vis}}^{(l)}
\left(\mathbf{h}_{i,l,t}\right)
+
\alpha_{i,l,t}^{\mathrm{code}}
\mathcal{B}_{\mathrm{code}}^{(l)}
\left(\mathbf{h}_{i,l,t}\right).
\end{equation}

By conditioning the branch weights on intermediate hidden representations, CAB dynamically coordinates visual grounding and code generation throughout the decoding process, enabling MoCA to accommodate diverse chart structures and evolving generation requirements.

\subsection{Model Training}

To train MoCA, we adopt a two-stage framework that progressively aligns structured reasoning with executable and visually faithful code generation. The first stage uses SFT to establish a stable mapping from chart images to structured reasoning trajectories and executable code. The second stage applies RL with execution and structure-aware feedback to further improve reasoning consistency, code validity, and visual fidelity. Together, the two stages provide complementary process-level and outcome-level supervision.

\subsubsection{Cross-Modal Snippet-of-Thought Alignment}

In the first stage, we establish a structured mapping from chart images to executable code. However, chart-to-code generation is inherently one-to-many: multiple implementations may produce visually equivalent charts while relying on different program structures. As a result, learning only from final code sequences provides limited supervision for the reasoning process that connects visual observations with code construction decisions.

We introduce a self-distilled SoT training strategy that decomposes the mapping process into concise intermediate steps. These trajectories provide structured supervision for connecting visual observations, chart structures, and code construction decisions. Our training data are derived from McD~\cite{jiang2025viscodex} and ChartCoder~\cite{zhao2025chartcoder}. We first obtain teacher-generated SoT trajectories and then apply self-distillation to produce supervision that is better aligned with MoCA's own generation distribution, reducing the discrepancy between externally generated reasoning patterns and the model's generation behavior. MoCA is optimized using standard autoregressive cross-entropy loss over the complete target sequence, including both the SoT trajectory and the final code.

\subsubsection{Multi-Dimensional Rewards Optimization}

In the second stage, we further optimize MoCA using RL with execution and structure-aware feedback. We adopt Group Relative Policy Optimization (GRPO)~\cite{shao2024deepseekmath}, which estimates relative advantages by comparing multiple responses generated for the same input, without requiring a separately learned critic model.

For each generated response $O=(S,C)$, where $S$ denotes the SoT trajectory and $C$ denotes the generated code, we compute five complementary reward signals. The format reward $R_{fmt}$ evaluates whether the response follows the required output format. The consistency reward $R_{con}$ measures the reasoning consistency between $S$ and $C$. The structural reward $R_{str}$ evaluates the reconstruction of key chart attributes through AST-based analysis. The execution reward $R_{exec}$ verifies whether $C$ executes successfully and produces a valid chart. The visual reward $R_{vis}$ measures the similarity between the rendered chart and the reference image. We aggregate the five signals into a scalar reward used by GRPO:
\begin{equation}
R(S, C) =
\underbrace{R_{fmt} + R_{con} + R_{str}}_{\text{Process-level}} + \underbrace{R_{exec} + R_{vis}}_{\text{Outcome-level}}
\end{equation}
The detailed configuration is provided in the supplementary material. This aggregation combines process-level guidance with outcome-level feedback, encouraging reasoning consistency, code executability, and visual fidelity.

\begin{table*}[t]
    \centering
    \small
    \caption{Performance of different models on ChartMimic, Plot2Code, and ChartX benchmarks.}
    \label{tab:model_evaluation_performance}
    \setlength{\tabcolsep}{2pt}
    \begin{tabular*}{\textwidth}{@{\extracolsep{\fill}} l ccc ccc c @{}}
        \toprule
        \textbf{Model} 
        & \multicolumn{3}{c}{\textbf{ChartMimic}} 
        & \multicolumn{3}{c}{\textbf{Plot2Code}} 
        & \textbf{ChartX} \\
        \cmidrule(lr){2-4} \cmidrule(lr){5-7} \cmidrule(l){8-8}
        & \textbf{Exec. Rate} & \textbf{Low-Level} & \textbf{High-Level} 
        & \textbf{Code Pass} & \textbf{Text Match} & \textbf{GPT-4o Score} 
        & \textbf{GPT Score} \\
        \midrule
        Full score & 100 & 100 & 100 & 100 & 100 & 10 & 5 \\
        \midrule
        \multicolumn{8}{c}{\textit{General-Domain Models}} \\
        \midrule
        GPT-4o-mini & \textbf{82.17} & 65.08 & 65.37 & \textbf{83.33} & 41.12 & \textbf{4.75} & 1.62 \\
        Qwen2.5-VL-72B   & 69.17 & 48.54 & 48.09 & 63.64 & \textbf{41.90} & 4.31 & 1.98 \\
        Qwen2.5-VL-32B   & 81.17 & 63.97 & \textbf{69.97} & 75.00 & 35.11 & 4.22 & \textbf{2.58} \\
        GLM-4.6V         & 48.33 & \textbf{73.03} & 43.67 & 33.33 & 18.50 & 2.20 & 2.47 \\
        InternVL3.5-38B  & 80.17 & 60.00 & 60.84 & 78.03 & 35.04 & 3.98 & 1.83 \\
        Qwen3-VL-30B-A3B & 76.67 & 61.67 & 61.78 & 59.85 & 27.85 & 3.25 & 1.94 \\
        \midrule
        \multicolumn{8}{c}{\textit{Chart-Specialized Models}} \\
        \midrule
        ChartLlama-13B   & 42.83 & 10.10 & 4.01 & 49.24 & 8.37 & 1.02 & 0.23 \\
        ChartCoder-7B    & 86.17 & 69.48 & 67.81 & 85.61 & 36.16 & 3.61 & 2.22 \\
        ChartVLM-L-14B   & 21.83 & 6.84 & 5.74 & 0.25 & 5.06 & 0.45 & 0.68 \\
        \textbf{MoCA} & \textbf{88.83} & \textbf{75.13} & \textbf{73.48} & \textbf{87.88} & \textbf{41.07} & 4.45 & \textbf{2.44} \\
        \bottomrule
    \end{tabular*}
\end{table*}

\section{Experiment}
We conduct comprehensive experiments on three benchmarks, including comparisons with existing models and detailed ablation analyses to validate the effectiveness of MoCA and its key components.
\vspace{-1pt}
\subsection{Experimental Settings}

\subsubsection{Model and Baselines}

We build MoCA upon Qwen2.5-VL-7B by replacing the FFN sublayer in each Transformer block with a CAB. The visual and code branches are initialized from the FFN parameters of Qwen2.5-VL-7B and Qwen2.5-Coder-7B, respectively. We compare MoCA with both general-domain and chart-specialized models. General-domain baselines include GPT-4o-mini~\cite{gpt4omini2024}, Qwen2.5-VL-72B/32B~\cite{qwen2025qwen25technicalreport}, GLM-4.6V~\cite{hong2025glm}, InternVL3.5-38B~\cite{wang2025internvl3}, and Qwen3-VL-30B-A3B~\cite{bai2025qwen3}. Chart-specialized baselines include ChartLlama-13B~\cite{han2023chartllama}, ChartCoder-7B~\cite{zhao2025chartcoder}, and ChartVLM-L-14B~\cite{xia2025chartx}.



\subsubsection{Evaluation Datasets and Metrics}

We evaluate MoCA on three widely used chart-to-code benchmarks. Following prior work~\cite{zhao2025chartcoder}, we use GPT-4o~\cite{hurst2024gpt} as the automatic evaluator for GPT-based visual similarity metrics. We apply a unified prompting protocol across all evaluated models and report Plot2Code results over the full evaluation set without filtering out non-executable outputs.

\subsubsection{Efficient Training and Inference Implementation}

We conduct SFT for one epoch on 8 NVIDIA H200 GPUs, using a learning rate of $1\times10^{-5}$ and a global batch size of 64. We conduct RL for three epochs on 32 NVIDIA H200 GPUs, using a learning rate of $1\times10^{-6}$ and generating eight responses per prompt. To support efficient execution of the dual-branch architecture, we implement custom QKV sharding and SwiGLU kernels and integrate them with vLLM.

\subsection{Overall Performance}

Table~\ref{tab:model_evaluation_performance} summarizes the performance of MoCA on three chart-to-code benchmarks. MoCA achieves the best result on four of the seven reported metrics and stays competitive on the rest. On ChartMimic, MoCA achieves the best performance across all three metrics, with an execution rate of 88.83, a low-level score of 75.13, and a high-level score of 73.48. These results exceed the strongest baseline on each metric by 2.66, 2.10, and 3.51 points, respectively. Notably, the improvements are consistent across both low-level and high-level criteria, whereas competing models often perform well on only one aspect. For example, GLM-4.6V achieves a low-level score of 73.03 but obtains only 43.67 on the high-level metric and an execution rate of 48.33, suggesting that it can reproduce local visual elements without reliably preserving global structure or generating executable code. On Plot2Code, MoCA achieves the highest Code Pass rate of 87.88, outperforming the next-best baseline by 2.27 points. Together with its leading execution rate on ChartMimic, this result indicates that MoCA’s advantage in executability is consistent across datasets. MoCA also remains competitive on semantic and perceptual metrics. Its Text Match score of 41.07 is only 0.83 points lower than the 41.90 achieved by the substantially larger Qwen2.5-VL-72B, while its ChartX GPT Score of 2.44 surpasses all chart-specialized baselines and remains competitive with larger general-domain models. Overall, MoCA jointly improves code executability and reconstruction fidelity while maintaining competitive holistic quality, demonstrating the effectiveness of coordinating visual grounding with code generation.

\subsection{Visualization of Arbitration Behavior}
\label{sec:arbitration_visualization}
To better understand the behavior of CAB, we visualize the branch weights produced by the Arbiter at multiple levels of granularity. 

\subsubsection{Token-Level}

\begin{figure}[t]
    \centering
    \includegraphics[width=\linewidth]{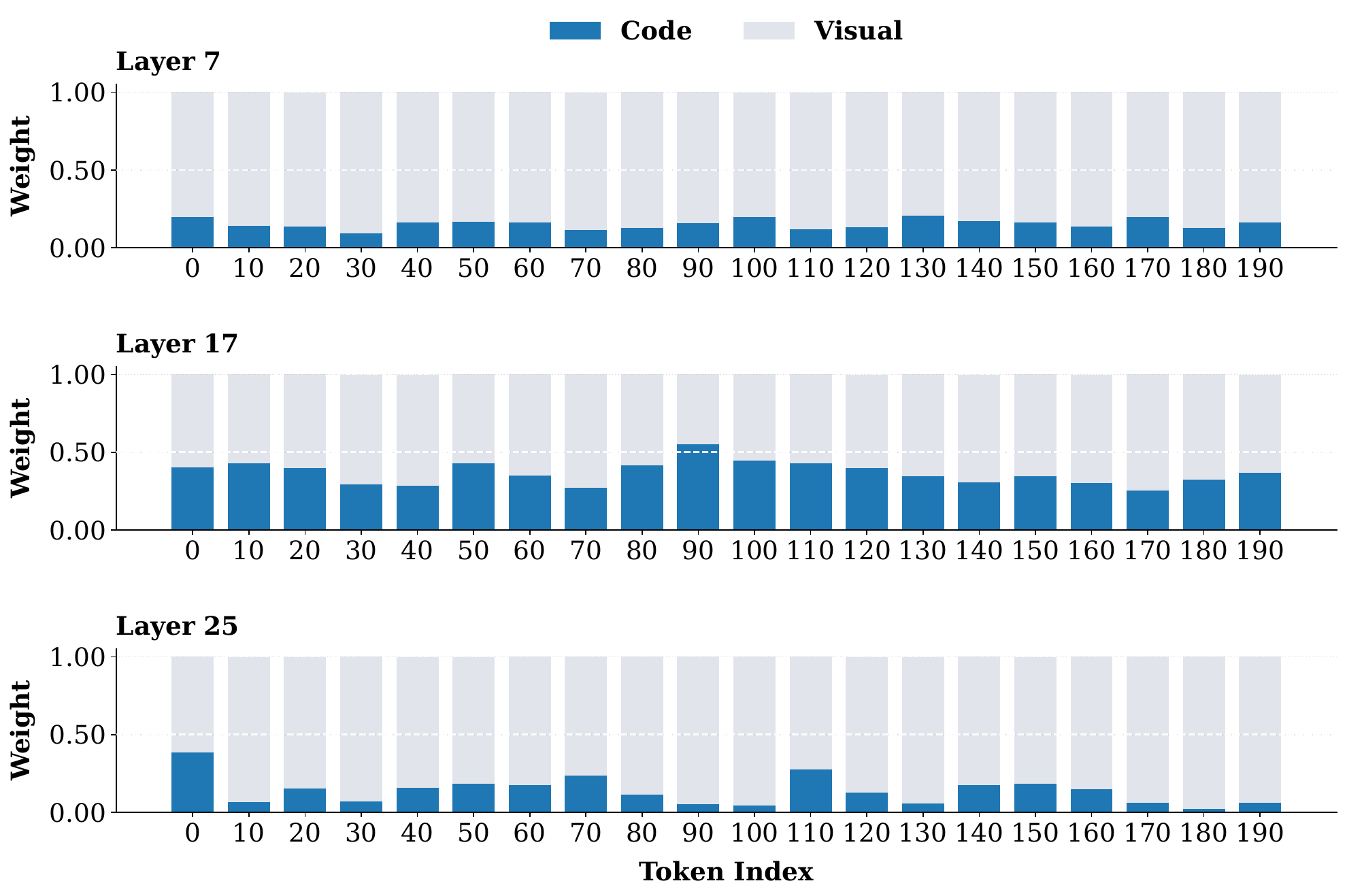}
    \caption{Visual and code branch weights across tokens at different layers (Layers 7, 17, and 25).}
    \label{fig:token_layer_arbitration}
\end{figure}

We first examine the branch weights across token positions in Figure~\ref{fig:token_layer_arbitration}. The weights vary across tokens within the same layer, with particularly noticeable variation in Layer 17. This indicates that the Arbiter does not apply a single sequence-level weighting throughout generation. Instead, it adjusts the relative contributions of the visual and code branches according to the evolving hidden representations, providing fine-grained coordination over the generation sequence.

\subsubsection{Layer-Level}

Figure~\ref{fig:token_layer_arbitration} also reveals distinct arbitration patterns across layers. Layer 7 is strongly dominated by the visual branch, whereas Layer 17 assigns a larger and more variable contribution to the code branch. At Layer 25, the allocation shifts back toward the visual branch. This non-monotonic yet structured pattern suggests stage-dependent coordination: shallow layers emphasize visual processing, intermediate layers increase the contribution of code generation, and deeper layers may return to visual refinement. Overall, these observations suggest that the Arbiter learns structured coordination patterns across network depth rather than applying uniform or arbitrary branch weights.

\subsubsection{Instance-Level}

\begin{figure}[t]
    \centering
    \includegraphics[width=\linewidth]{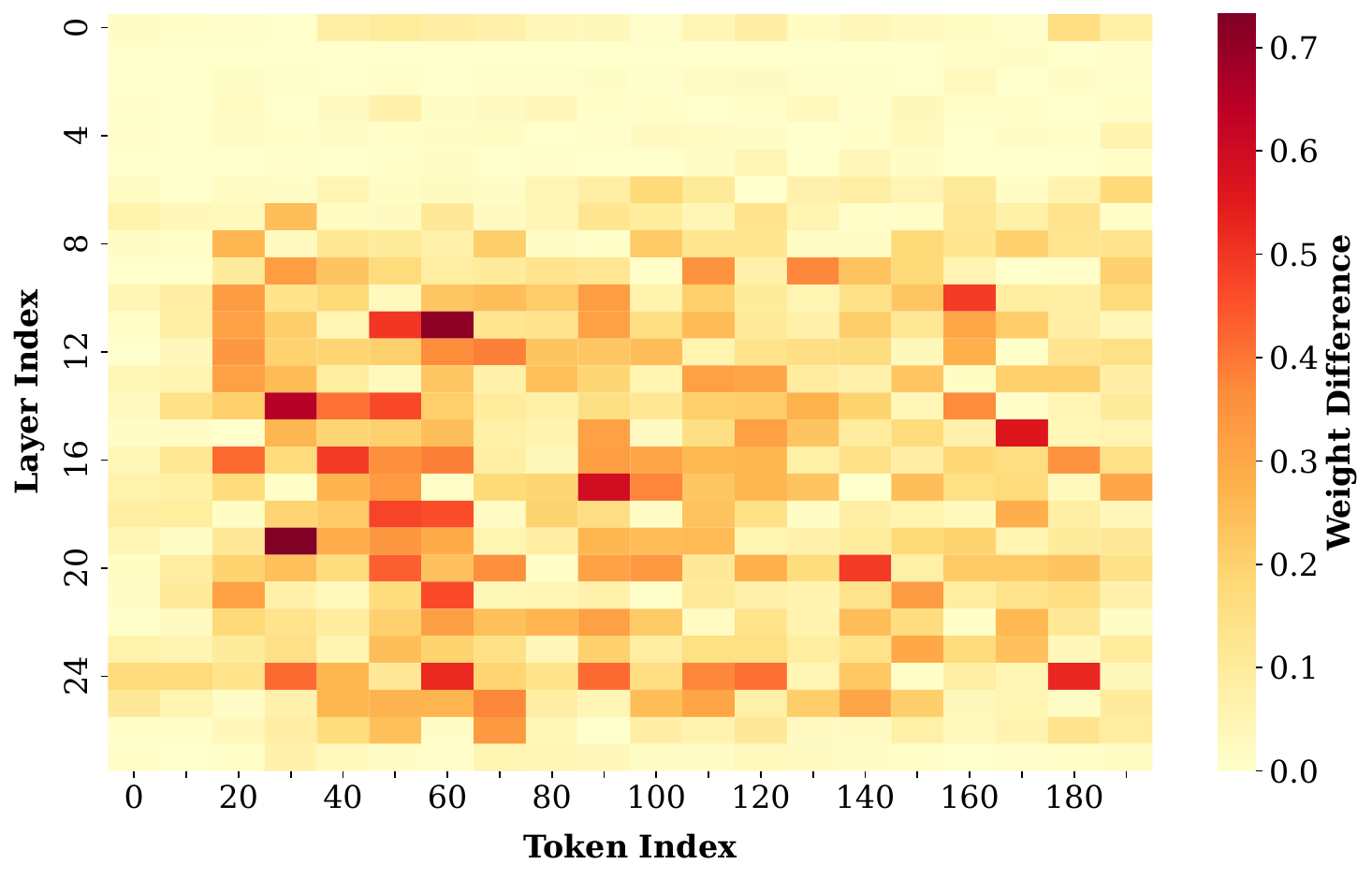}
    \caption{Absolute differences in branch weights between two input samples across layers and tokens.}
    \label{fig:instance_arbitration}
\end{figure}

Figure~\ref{fig:instance_arbitration} shows the absolute differences in branch weights between two randomly selected Plot2Code inputs. The differences are relatively small in the early layers but become more pronounced across several middle and deeper layers, with localized variations across token positions. This provides qualitative evidence that the Arbiter produces instance-specific arbitration patterns rather than applying the same allocation to every input. Since the visualization contains only one pair of samples, we interpret it as an illustrative case rather than a comprehensive measure of input adaptivity.

\subsection{Ablation Study}

\begin{table}[t]
    \centering
    \small
    \caption{Ablation of SFT and RL for MoCA on Plot2Code.}
    \label{tab:ablation_sft_rl}
    \begin{tabular}{ccccc}
        \toprule
        \multicolumn{2}{c}{\textbf{Training}} 
        & \multicolumn{3}{c}{\textbf{Plot2Code}} \\
        \cmidrule(lr){1-2} \cmidrule(lr){3-5}
        \textbf{SFT} & \textbf{RL} 
        & \textbf{Code Pass}
        & \textbf{Text Match}
        & \textbf{GPT-4o Score} \\
        \midrule
        No  & No  & -- & -- & -- \\
        Yes & No  & 83.33 & 34.11 & 3.66 \\
        No  & Yes & -- & -- & -- \\
        Yes & Yes & \textbf{87.88} & \textbf{41.07} & \textbf{4.45} \\
        \bottomrule
    \end{tabular}
\end{table}

\subsubsection{Effectiveness of Two-Stage Model Training}
We analyze the complementary roles of SFT and RL in training MoCA. As shown in Table~\ref{tab:ablation_sft_rl}, MoCA without task-specific training and MoCA trained with RL alone fail to produce valid structured outputs, and their results are therefore denoted by ``--''. In contrast, SFT establishes a stable mapping from visual inputs to executable code, achieving a Code Pass rate of 83.33, a Text Match score of 34.11, and a GPT-4o Score of 3.66 on Plot2Code. Building on the SFT model, RL with the multi-dimensional reward function further improves the Code Pass rate to 87.88, the Text Match score to 41.07, and the GPT-4o Score to 4.45, corresponding to absolute gains of 4.55, 6.96, and 0.79 points, respectively. These results indicate that SFT establishes reliable code generation, while RL further improves code executability, structural fidelity, and visual similarity.

\begin{table}[t]
    \centering
    \small
    \caption{Ablation results on Plot2Code evaluating the effects of two-stage training, branch initialization, static parameter merging, and input-conditioned arbitration.}
    \label{tab:merge_ablation}
    \setlength{\tabcolsep}{5pt}
    \begin{tabular}{llccc}
        \toprule
        \multirow{3}{*}{\textbf{Method}}
        & \multirow{3}{*}{\textbf{Training}}
        & \multicolumn{3}{c}{\textbf{Plot2Code}} \\
        \cmidrule(lr){3-5}
        &
        & \makecell{\textbf{Code}\\\textbf{Pass}}
        & \makecell{\textbf{Text}\\\textbf{Match}}
        & \makecell{\textbf{GPT-4o}\\\textbf{Score}} \\
        \midrule

        \multirow{3}{*}{\makecell[l]{Base\\\footnotesize(Qwen2.5-VL-7B)}}
        & None     & 71.97 & 34.85 & 3.47 \\
        & SFT      & 74.24 & 31.40 & 3.30 \\
        & SFT + RL & 84.10 & 36.11 & 3.87 \\
        \midrule

        \multirow{2}{*}{\makecell[l]{Visual + Visual\\\footnotesize(w/o code)}}
        & SFT      & 79.92 & 33.86 & 3.55 \\
        & SFT + RL & 85.26 & 37.48 & 3.94 \\
        \midrule

        \multirow{2}{*}{\makecell[l]{Static merging\\\footnotesize(VisCodex)}}
        & SFT      & 69.70 & 27.38 & 2.79 \\
        & SFT + RL & 70.45 & 31.86 & 3.15 \\
        \midrule

        \multirow{2}{*}{\makecell[l]{Fixed branch weights\\\footnotesize(w/o Arbiter)}}
        & SFT      & 84.32 & 35.20 & 3.21 \\
        & SFT + RL & 86.74 & 39.82 & 4.26 \\
        \midrule

        \textbf{MoCA (Ours)}
        & \textbf{SFT + RL}
        & \textbf{87.88}
        & \textbf{41.07}
        & \textbf{4.45} \\

        \bottomrule
    \end{tabular}
\end{table}

\begin{figure*}[!t]
    \centering
    \includegraphics[width=0.85\textwidth]{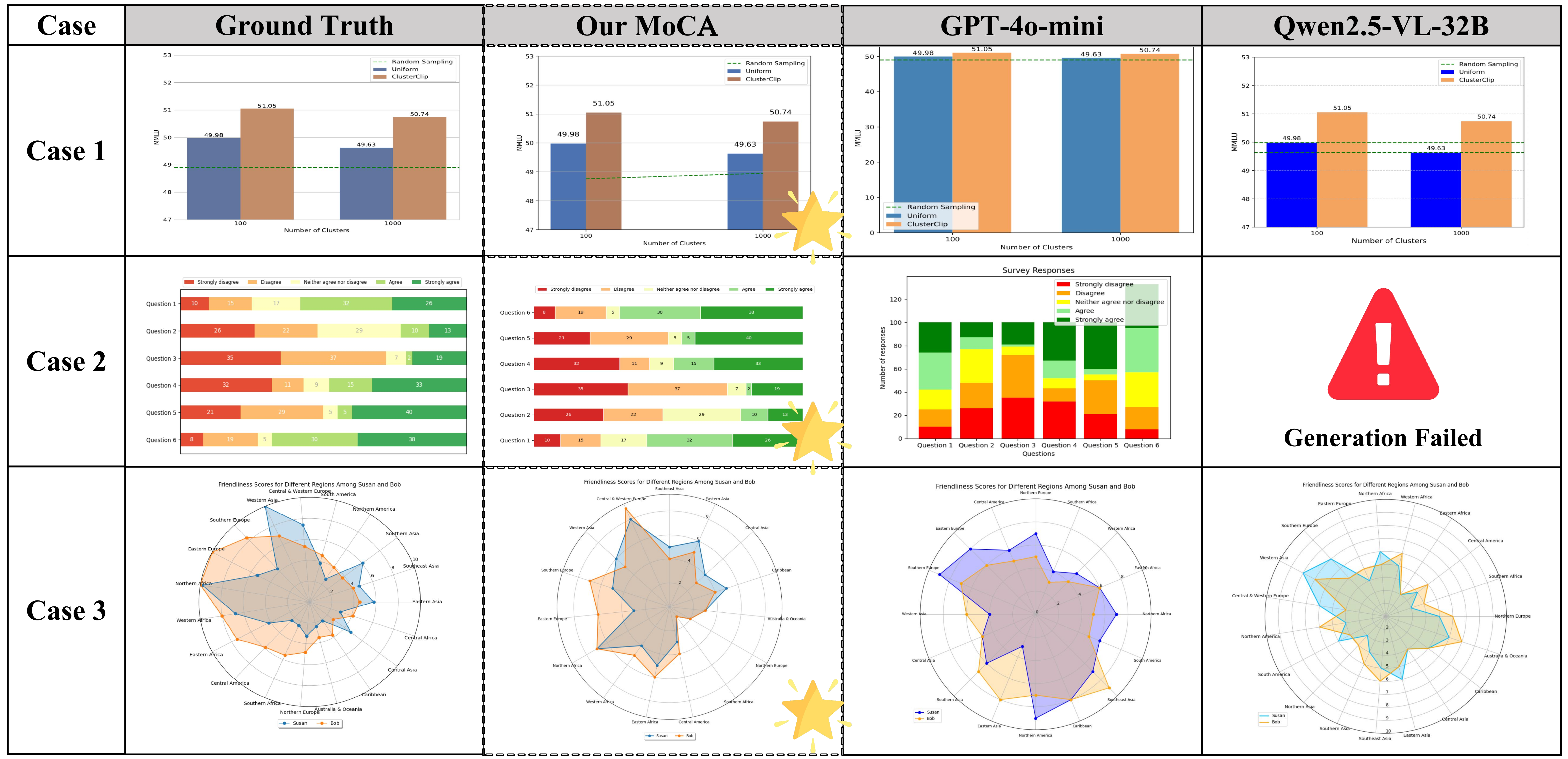}
    \caption{Comparison of MoCA with other LLM backbones.}
    \label{case_study}
\end{figure*}

\subsubsection{Effect of Branch Initialization and Arbitration}
We investigate the sources of MoCA's improvement, including task-specific training, model capacity, branch initialization, and input-conditioned arbitration. Table~\ref{tab:merge_ablation} presents the results on Plot2Code. We first compare MoCA with the original Qwen2.5-VL-7B backbone. After SFT and RL, the backbone achieves a Code Pass rate of 84.10, a Text Match score of 36.11, and a GPT-4o Score of 3.87. Although task-specific training substantially improves its performance, it remains inferior to MoCA across all three metrics, indicating that training alone is insufficient to effectively coordinate visual understanding and code generation.

We next examine the effect of complementary branch initialization. We construct a Visual + Visual variant that retains the same two-branch architecture as MoCA but initializes both branches from Qwen2.5-VL-7B. This variant achieves scores of 85.26, 37.48, and 3.94, whereas MoCA improves upon them by 2.62, 3.59, and 0.51 points, respectively. These results show that the gains arise from combining complementary visual and coding expertise rather than merely increasing the number of parameters. We also compare MoCA with static parameter merging. The VisCodex-based baseline achieves Code Pass, Text Match, and GPT-4o scores of 70.45, 31.86, and 3.15, respectively, suggesting that directly merging visual and coding parameters is insufficient for integrating heterogeneous capabilities.

Finally, we evaluate the role of the Arbiter using a fixed-weight variant that retains the same branches and training configuration as MoCA but sets $\alpha^{\mathrm{vis}}=\alpha^{\mathrm{code}}=0.5$ for every token and layer. This variant achieves scores of 86.74, 39.82, and 4.26, respectively. MoCA's consistent improvements over fixed weighting demonstrate the effectiveness of input-conditioned arbitration. Overall, the ablation results show that MoCA's gains cannot be attributed to task-specific training or increased model size alone, but instead arise from the joint contributions of complementary branch initialization and input-conditioned arbitration.

\vspace{-2pt}
\subsubsection{Impact of Self-distilled SoT Data}
We evaluate the effect of self-distillation by comparing the original teacher-generated SoT trajectories with self-distilled trajectories under the same SFT configuration. On Plot2Code, replacing the original SoT data with self-distilled SoT data increases the Code Pass rate from 80.30 to 83.33. This result suggests that self-distillation better aligns the training trajectories with the model's own output distribution, thereby narrowing the training--inference gap and improving generalization.

\vspace{-1pt}
\section{Case Study}
To qualitatively evaluate our method, we present representative outputs from different models in Figure~\ref{case_study}. For the bar chart example, GPT-4o-mini fails to produce correct y-axis scales and misplaces the legend, which results in incomplete semantics. In contrast, MoCA correctly reconstructs the data distribution, dashed reference lines, and legend, while maintaining consistent color mapping. The stacked chart is more challenging due to the larger number of categories, complex color assignments, and the need for accurate value accumulation. Qwen2.5-VL-32B fails to generate a valid chart due to code errors. GPT-4o-mini produces a vertical layout instead of the original horizontal stacking, breaking the intended structure. MoCA preserves the stacking direction and better matches both color assignments and value proportions. For more complex structures such as the radar chart, MoCA preserves the overall topology and maintains correct alignment. Overall, MoCA produces more stable results in scale recovery, structural preservation, and multi-component coordination, suggesting more effective interaction between visual understanding and code generation.
\vspace{-2pt}
\section{Conclusion}

\vspace{-1pt}
We present MoCA, a chart-to-code framework that coordinates a visual branch and a code branch to generate executable code with faithful visual reconstruction. Through CAB, a lightweight Arbiter dynamically adjusts the relative contributions of the two branches for each token at every layer based on intermediate hidden representations. Combined with SFT on self-distilled SoT trajectories and RL with multi-dimensional rewards, MoCA achieves strong code executability and competitive visual fidelity across three benchmarks. Ablation results show that these improvements cannot be explained by additional training or model size alone, highlighting the importance of complementary branch initialization and input-conditioned arbitration over static parameter merging and fixed branch weighting. Future work will investigate more parameter-efficient fusion paradigms, aiming to achieve effective visual-code integration without increasing model parameters.

\bibliography{reference}

\end{document}